\pdfoutput=1
\documentclass[11pt]{article}

\usepackage[final]{acl}

\usepackage{times}
\usepackage{latexsym}

\usepackage[T1]{fontenc}
\usepackage[utf8]{inputenc}

\usepackage{microtype}

\usepackage{inconsolata}
\usepackage{amsmath}
\usepackage{amssymb}
\usepackage{booktabs}
\usepackage{graphicx}
\usepackage{algorithm}
\usepackage{algorithmicx}
\usepackage{algpseudocode}
\usepackage[most]{tcolorbox}
\usepackage{bbding}
\usepackage{multirow}
\usepackage{multicol}
\usepackage{rotating}
\usepackage[inkscapelatex=false]{svg}
\usepackage[font=small,labelfont=bf]{subcaption}
\usepackage{anyfontsize}
\usepackage{tikz}
\usepackage{pgfplots}
\pgfplotsset{compat=1.18}
\usepackage{scalefnt}

\usepackage{graphicx}

\usepackage{cleveref}
\crefname{section}{\S}{\S\S}
\Crefname{section}{\S}{\S\S}
\crefname{table}{Tab.}{Tabs.}
\crefname{figure}{Fig.}{Figs.}
\crefname{algorithm}{Alg.}{}
\crefname{appendix}{App.}{Apps.}
\crefname{lemma}{Lemma}{}
\Crefname{theorem}{Theorem}{}
\crefname{proposition}{Proposition}{}
\crefname{hypothesis}{Hypothesis}{}
\crefname{deduction}{Deduction}{}
\crefname{intuition}{\textbf{Intuition}}{\textbf{Intuitions}}
\crefname{observation}{\textbf{Observation}}{\textbf{Observations}}
\crefname{finding}{\textbf{Finding}}{\textbf{Findings}}
\crefname{cor}{Corollary}{}
\crefname{align}{}{}
\crefname{equation}{}{}

\usepackage{enumitem}
\setlist{leftmargin=*}

\definecolor{ETHBlue}{RGB}{33,92,175}	% blue 
\definecolor{ETHGreen}{RGB}{98,115,19}		% green
\definecolor{ETHPurpleDark}{RGB}{140,10,89}	% purple
\definecolor{ETHPurple}{RGB}{163,7,116}	% purple
\definecolor{ETHPurpleLight}{RGB}{220, 158, 201}	% purple
\definecolor{ETHGray}{RGB}{111,111,111}	% gray
\definecolor{ETHRed}{RGB}{183,53,45}	% red
\definecolor{ETHPetrol}{RGB}{0,120,148}	% green/blue 
\definecolor{ETHBronze}{RGB}{142,103,19}	% bronze

\newtoggle{color-macro}
\settoggle{color-macro}{false} % set to false to disable color macros

\iftoggle{color-macro}{
\colorlet{MacroColor}{ETHPetrol}
}{
\colorlet{MacroColor}{black}
}

\newcommand{\appleimage}{\protect\includegraphics[height=1.4em]{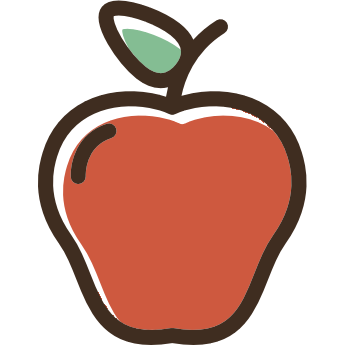}}
\newcommand{\bananaimage}{\protect\includegraphics[height=1.4em]{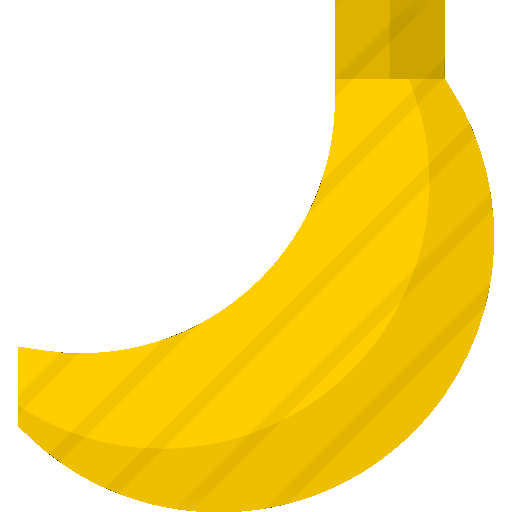}}

\title{Can Vision-Language Models Solve Visual Math Equations?}

\author{
    \textbf{Monjoy Narayan Choudhury\textsuperscript{*}\textsuperscript{1}}, 
    \textbf{Junling Wang\textsuperscript{*}\textsuperscript{2}}, 
    \textbf{Yifan Hou\textsuperscript{2}}, 
    \textbf{Mrinmaya Sachan\textsuperscript{2}},
\\
 \textsuperscript{1}IIIT Bangalore,
 \textsuperscript{2}ETH Z\"{u}rich,
\\
    \texttt{\textsuperscript{1}\href{mailto:monjoy.choudhury@iiitb.ac.in}{monjoy.choudhury@iiitb.ac.in},}\\
\textsuperscript{2}$\{$
    \texttt{{\href{mailto:junling.wang@inf.ethz.ch}{junling.wang},}
    \texttt{\href{mailto:yifan.hou@inf.ethz.ch}{yifan.hou},}
    \texttt{\href{mailto:mrinmaya.sachan@inf.ethz.ch}{mrinmaya.sachan}}
    $\}$\texttt{@inf.ethz.ch}}
}
\begin{document}
\maketitle
\def\thefootnote{*}\footnotetext{Equal contribution.}\def\thefootnote{\arabic{footnote}}

\begin{abstract}
Despite strong performance in visual understanding and language-based reasoning, Vision-Language Models (VLMs) struggle with tasks requiring integrated perception and symbolic computation. 
We study this limitation through \textit{visual equation solving}, where mathematical equations are embedded in images, variables are represented by object icons, and coefficients must be inferred by counting. 
While VLMs perform well on textual equations, they fail on visually grounded counterparts. To understand this gap, we decompose the task into coefficient counting and variable recognition, and find that counting is the primary bottleneck, even when recognition is accurate. 
We also observe that composing recognition and reasoning introduces additional errors, highlighting challenges in multi-step visual reasoning. 
Finally, as equation complexity increases, symbolic reasoning itself becomes a limiting factor. 
These findings reveal key weaknesses in current VLMs and point toward future improvements in visually grounded mathematical reasoning.\footnote{Our \href{https://github.com/eth-lre/MathEval}{code} and \href{https://huggingface.co/datasets/monjoychoudhury29/Visual-Math-Eval}{data} are publicly available.}

\end{abstract}

\section{Introduction}
Vision-Language Models (VLMs) have become the dominant architecture for multimodal learning, powering applications such as visual question answering~\citep{ghosal-etal-2023-language}, image captioning~\citep{yang-etal-2023-vilm}, and multimodal reasoning~\citep{li-etal-2024-enhancing-advanced}. 
As agentic AI systems gain traction, VLMs are increasingly expected to function as general-purpose perception-and-reasoning modules for intelligent agents~\citep{li-etal-2024-enhancing-advanced,li-etal-2024-topviewrs}. While recent models demonstrate strong capabilities in both visual understanding and language-based reasoning, truly agentic behavior demands deeper integration, particularly in tasks involving grounded mathematical reasoning~\cite{shi-etal-2024-math}.

In this work, we investigate this integration through the lens of a seemingly simple but revealing task: \textit{visual equation solving}. Given an image containing a system of equations where variables are depicted as object icons (e.g., $\appleimage\appleimage + \bananaimage = 10$), the goal is to infer coefficients by counting icons and solve the equation accordingly. While this task appears tractable for models proficient in both visual and symbolic reasoning, our results show that even the strongest VLMs fail to solve such problems reliably.
\textit{Why do VLMs struggle with visual equation solving?} To answer this, we decompose the task into two core components: \textit{symbolic equation solving} and \textit{visual recognition}.

We begin by testing symbolic reasoning in isolation. When equations are presented in plain text in the image, VLMs solve them almost perfectly, confirming their mathematical reasoning and OCR capabilities. Next, we evaluate whether variable recognition is the bottleneck. Models are able to correctly identify object-based variables with high accuracy, suggesting recognition alone is not the issue.
We then turn to coefficient estimation, counting the number of object instances. In hybrid settings where variables are icons and coefficients are numerals, or where both are visual, performance drops significantly. Direct evaluation of object counting further confirms that this is the key bottleneck: \textit{VLMs often fail to infer quantities from repeated visual elements}.

Beyond counting, we observe that performance degrades further when multiple abilities, such as recognition and reasoning, must be composed. For instance, even when a model can recognize variables and solve symbolic equations separately, solving equations with icon-based variables and numeric coefficients proves difficult. This highlights compositional reasoning as another major challenge for current VLMs.
Finally, we evaluate systems of equations with three variables. Even when equations are presented symbolically, performance drops sharply, indicating that VLMs' mathematical reasoning is itself limited when faced with more complex problem structures.

Taken together, our findings reveal key limitations in current VLMs’ ability to integrate perception and symbolic reasoning. In particular, visual counting and ability composition emerge as core bottlenecks, alongside limited generalization in symbolic math reasoning for complex tasks.

\section{Preparation}
We design a controlled evaluation setup to analyze VLMs’ ability to perform visual equation solving. This section describes our data generation process and the experimental settings used for the following model evaluation.

\subsection{Data}
We construct synthetic visual math problems based on systems of linear equations, where variables are depicted as object icons and coefficients must be inferred from visual repetition. Each experiment is conducted on a set of 1,000 constructed examples and run once per model-setting configuration.

\paragraph{Equation Generation.}
We generate solvable systems of linear equations with unique integer solutions using matrix algebra, ensuring invertibility. To control visual complexity, coefficients are restricted to positive integers no greater than 10, limiting the number of repeated icons per image. All equations involve only addition, avoiding negative or fractional values. This setup ensures consistency and interpretability across all samples.

\begin{figure}[!ht]
    \centering
    %\includesvg[inkscapelatex=false, width=0.98\linewidth]{./figures/object-type.svg}
    \includegraphics[width=0.98\linewidth]{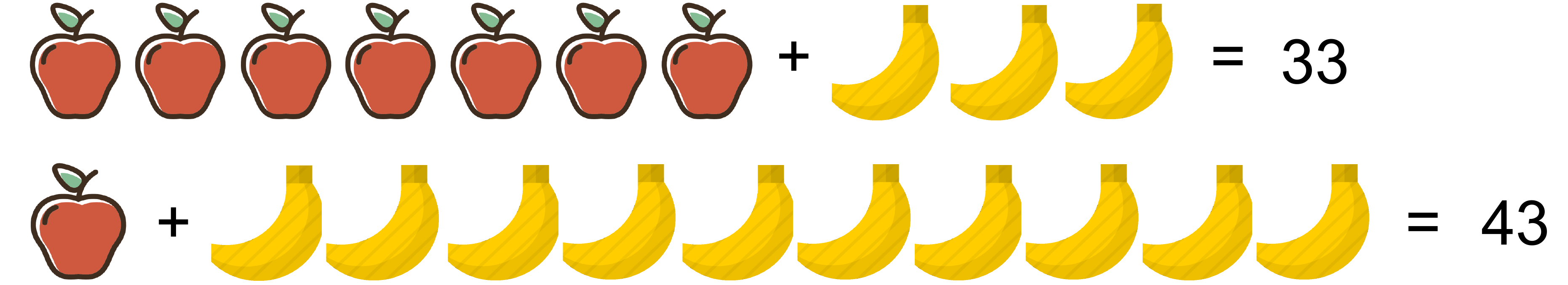}
    \caption{An example of our generated visual equations (i.e., systems of 2 linear equations with 2-variables).}
    %\vspace{-.2cm}
    \label{fig:icon_all}
\end{figure}

\paragraph{Image Construction.}
To visually represent equations, we map each variable to an icon selected from a curated set of 28 object types in the IconQA dataset~\citep{iconqa}, including items such as apples, bananas, flowers, and footballs. The coefficient of each variable is represented by repeating the corresponding icon the appropriate number of times. This creates visually grounded equations that require both recognition and symbolic reasoning. An example is shown in \cref{fig:icon_all}, and the full list of icons is provided in \cref{app:details:icon}.

\subsection{Settings}

\paragraph{Model List.}
We evaluate both proprietary and open-source VLMs. The former include GPT-4o~\citep{gpt4o} and Gemini 2.0 Flash~\citep{gemini}, accessed via API. The latter consist of four models from the QwenVL-2.5 family~\citep{qwenVL2023}, ranging from 3B to 72B parameters. To ensure fairness, all models are evaluated without batching, avoiding potential artifacts from cached context or batch-level optimizations. More details about the model can be found in \cref{app:details:settings}.

\paragraph{Prompting Strategy.}
We apply two prompting strategies: direct zero-shot prompting (Direct) and two-step chain-of-thought (CoT) prompting. In the CoT setting, the model is first asked to extract the equation in free-form, then solve it in a second step. This setup encourages intermediate reasoning before committing to an answer. Both strategies are applied consistently across all models. Prompt templates and examples are provided in \cref{app:details:prompt}.

\paragraph{Metrics.}
We evaluate accuracy by exact matching between the model-predicted variable values and the ground truth. 
We expect models to correctly associate each object type with its corresponding value and solve the equation. 
%Considering that all equations are guaranteed to have unique integer solutions, fractional predictions are considered incorrect. This setup assumes VLMs can apply basic world knowledge, i.e., that objects are countable and discrete.

\section{Evaluation}
We evaluate the mathematical reasoning capabilities of VLMs through the task of visual equation solving. Specifically, we investigate two research questions:
(1) Can VLMs solve equations when they are visually grounded?
(2) If not, what specific limitations hinder their performance?

\subsection{Can VLMs Solve Equations?}
% visual: no, symbolic: yes
We begin our evaluation on solving systems of linear equations in two formats: (1) a fully visual format, where both variables and coefficients are depicted visually (\cref{fig:icon_all}), and (2) a symbolic format, where equations are rendered as text within the image (\cref{fig:icon_none}). This comparison can isolate the impact of visual understanding on performance.

\subsubsection{Visual Equation}
\paragraph{Experiment Preparation.}
We use a default setting of two-variable linear equations with integer solutions. In each equation, variables are represented by object icons, and coefficients are conveyed by the number of repeated instances of each icon. This setup tests whether VLMs can integrate visual perception and symbolic reasoning.

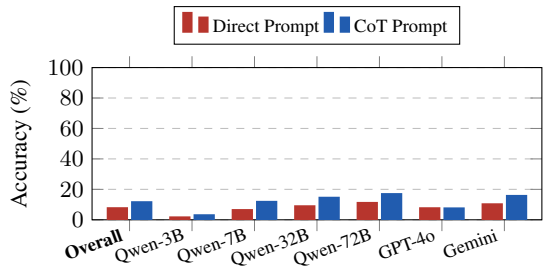
\begin{figure}[!ht]
%\begin{wrapfigure}[12]{r}{0.6\textwidth}
    \centering
    \scalefont{0.8}
    \begin{tikzpicture}
    \begin{axis}[
        ybar,
        bar width=0.25cm,
        width=7.5cm, height=3.6cm,
        enlarge x limits=0.1,
        ylabel={Accuracy (\%)},
        xlabel near ticks,
        ylabel near ticks,
        ymin=0, ymax=100,
        xtick=data,
        %bar shift=0
        xticklabel style={rotate=20, anchor=east, align=left, font=\scriptsize},
        symbolic x coords={\textbf{Overall}, Qwen-3B, Qwen-7B, Qwen-32B, Qwen-72B, GPT-4o, Gemini},
        legend style={at={(0.5,1.15)}, anchor=south, legend columns=3, nodes={scale=0.8, transform shape}},
        ymajorgrids=true,
        grid style=dashed,
    ]

    % Real
    \addplot+[line width=0.3mm, color=ETHRed] coordinates {
        (\textbf{Overall}, 7.53)
        (Qwen-3B, 1.50)
        (Qwen-7B, 6.30)
        (Qwen-32B, 8.80)
        (Qwen-72B, 11.00)
        (GPT-4o, 7.50)
        (Gemini, 10.10)
    };
    \addlegendentry{Direct Prompt}
    % Real
    \addplot+[line width=0.3mm, color=ETHBlue] coordinates {
        (\textbf{Overall}, 11.47)
        (Qwen-3B, 2.90)
        (Qwen-7B, 11.70)
        (Qwen-32B, 14.4)
        (Qwen-72B, 16.8)
        (GPT-4o, 7.40)
        (Gemini, 15.60)
    };
    \addlegendentry{CoT Prompt}
    \end{axis}
    \end{tikzpicture}
    \caption{Performance of VLMs on visual equation solving. Results show that all models consistently fail to solve the equations correctly across both settings.}
    \label{fig:results:icon_all}
\end{figure}
%\end{wrapfigure}
\paragraph{Results and Analysis.} 
As shown in \cref{fig:results:icon_all}, all evaluated models, both proprietary and open-source, consistently fail to solve equations in visual form (overall accuracy $<$ 12\%), despite their strong performance on other math and reasoning benchmarks. To rule out flaws in the evaluation setup, we include qualitative model outputs in \cref{app:results:case_study}. These results raise a key question: Is the failure due to a lack of symbolic math reasoning, or a difficulty in interpreting equations visually?

\subsubsection{Symbolic Equation}
\begin{figure}[!ht]
    \centering
    \includegraphics[width=\columnwidth]{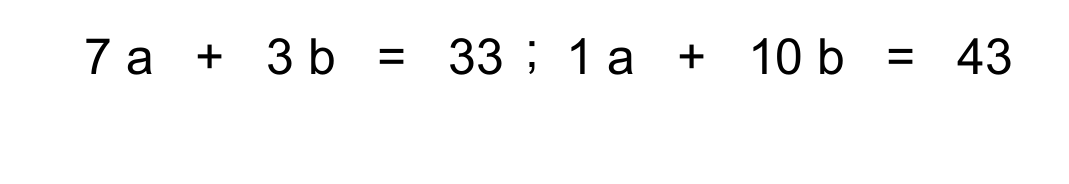}
    \caption{An example of a system of linear equations represented in symbolic (textual) form.}
    \label{fig:icon_none}
\end{figure}

\paragraph{Experiment Preparation.}
To isolate symbolic reasoning ability, we present the same equations in textual form within images (\cref{fig:icon_none}). If models succeed here, it would suggest that the core issue lies in interpreting the visual input, not solving the equations themselves.

\begin{figure}[!ht]
%\begin{wrapfigure}[12]{r}{0.6\textwidth}
    \centering
    \scalefont{0.8}
    \begin{tikzpicture}
    \begin{axis}[
        ybar,
        bar width=0.25cm,
        width=7.5cm, height=3.6cm,
        enlarge x limits=0.1,
        ylabel={Accuracy (\%)},
        xlabel near ticks,
        ylabel near ticks,
        ymin=0, ymax=100,
        xtick=data,
        %bar shift=0
        xticklabel style={rotate=20, anchor=east, align=left, font=\scriptsize},
        symbolic x coords={\textbf{Overall}, Qwen-3B, Qwen-7B, Qwen-32B, Qwen-72B, GPT-4o, Gemini},
        legend style={at={(0.5,1.15)}, anchor=south, legend columns=3, nodes={scale=0.8, transform shape}},
        ymajorgrids=true,
        grid style=dashed,
    ]

    % Real
    \addplot+[line width=0.3mm, color=ETHRed] coordinates {
        (\textbf{Overall}, 30.10)
        (Qwen-3B, 2.40)
        (Qwen-7B, 18.30)
        (Qwen-32B, 44.10)
        (Qwen-72B, 34.70)
        (GPT-4o, 28.80)
        (Gemini, 52.30)
    };
    \addlegendentry{Direct Prompt}
    % Real
    \addplot+[line width=0.3mm, color=ETHBlue] coordinates {
        (\textbf{Overall}, 98.42)
        (Qwen-3B, 97.90)
        (Qwen-7B, 97.90)
        (Qwen-32B, 98.90)
        (Qwen-72B, 98.60)
        (GPT-4o, 98.40)
        (Gemini, 98.80)
    };
    \addlegendentry{CoT Prompt}
    \end{axis}
    \end{tikzpicture}
    \caption{Performance of VLMs on symbolic equation solving. Results show that all models could solve the equations perfectly across settings.}
    \label{fig:results:icon_none}
\end{figure}
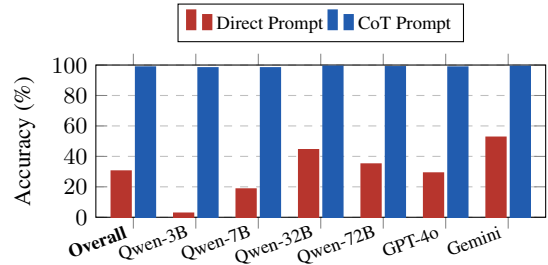
%\end{wrapfigure}
\paragraph{Results and Analysis.}
\cref{fig:results:icon_none} shows that all models, including the smallest Qwen-3B, achieve near-perfect accuracy on symbolic equations (accuracy $>$ 97\% with the CoT prompting). This confirms two things: (1) VLMs possess the required mathematical reasoning capabilities, and (2) they have strong OCR skills for extracting text from images. These findings indicate that the failure in the visual setting stems from difficulties in interpreting and grounding visual equations.

\subsection{Visual-Symbolic Gap Analysis}
To understand the source of the performance gap between visual and symbolic settings, we decompose visual equation solving into two core sub-skills:
(1) recognizing variables from icons, and
(2) estimating coefficients by counting repeated visual instances.
This allows us to evaluate whether recognition or counting is the main bottleneck, or whether it arises from composing the two abilities.

\subsubsection{Coefficient Counting}

\begin{figure}[!ht]
    \centering
    \includegraphics[width=\columnwidth]{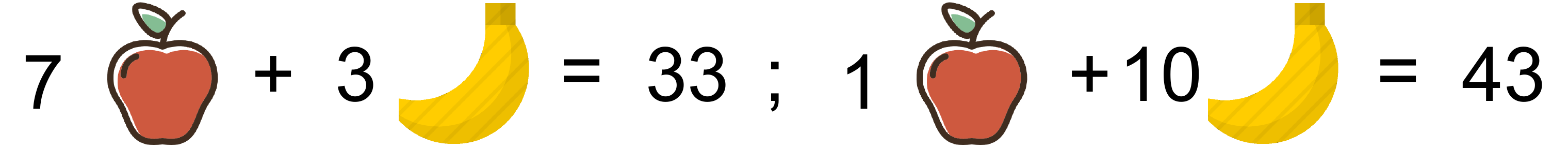}
    \caption{An example of our generated visual-symbolic equation, where the variable is denoted by icon but the coefficient is represented by symbolic number.}
    \label{fig:partial-variant}
\end{figure}

\paragraph{Experiment Preparation.}
We design a hybrid variant called visual-symbolic equations (\cref{fig:partial-variant}), where variables are represented as icons, but coefficients are given as numeric text. This setting removes the need for counting while preserving the need for icon recognition and symbolic reasoning.

\begin{figure}[!ht]
%\begin{wrapfigure}[12]{r}{0.6\textwidth}
    \centering
    \scalefont{0.8}
    \begin{tikzpicture}
    \begin{axis}[
        ybar,
        bar width=0.25cm,
        width=7.5cm, height=3.6cm,
        enlarge x limits=0.1,
        ylabel={Accuracy (\%)},
        xlabel near ticks,
        ylabel near ticks,
        ymin=0, ymax=100,
        xtick=data,
        %bar shift=0
        xticklabel style={rotate=20, anchor=east, align=left, font=\scriptsize},
        symbolic x coords={\textbf{Overall}, Qwen-3B, Qwen-7B, Qwen-32B, Qwen-72B, GPT-4o, Gemini},
        legend style={at={(0.5,1.15)}, anchor=south, legend columns=3, nodes={scale=0.8, transform shape}},
        ymajorgrids=true,
        grid style=dashed,
    ]

    % Real
    \addplot+[line width=0.3mm, color=ETHRed] coordinates {
        (\textbf{Overall}, 7.92)
        (Qwen-3B, 0.9)
        (Qwen-7B, 1.6)
        (Qwen-32B, 13.1)
        (Qwen-72B, 20.1)
        (GPT-4o, 4.2)
        (Gemini, 7.6)
    };
    \addlegendentry{Direct Prompt}
    % Real
    \addplot+[line width=0.3mm, color=ETHBlue] coordinates {
        (\textbf{Overall}, 64.45)
        (Qwen-3B, 34.7)
        (Qwen-7B, 56)
        (Qwen-32B, 53.7)
        (Qwen-72B, 86.6)
        (GPT-4o, 84.9)
        (Gemini, 70.8)
    };
    \addlegendentry{CoT Prompt}
    \end{axis}
    \end{tikzpicture}
    \caption{Performance of VLMs on visual-symbolic equation solving, where the coefficients are represented by symbolic numbers and variables are denoted by icons. Results show that all models could solve most systems of equations correctly.}
    \label{fig:results:icon_partial}
\end{figure}
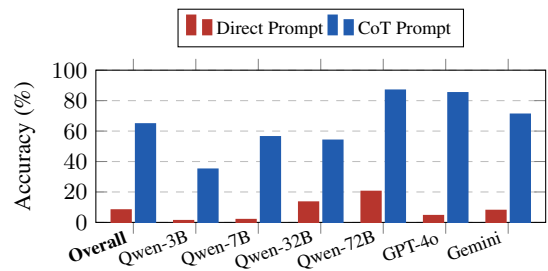
%\end{wrapfigure}

\paragraph{Results and Analysis.}
As shown in \cref{fig:results:icon_partial}, VLMs perform better in this setting than in the fully visual case (with overall accuracy as 64.45\%), suggesting that coefficient counting is a major obstacle. To further confirm this, we directly evaluate models on isolated counting tasks (see \cref{app:results:counting}). These results clearly identify counting as a primary bottleneck in visual equation solving.

\subsubsection{Variable Recognition}
\paragraph{Experiment Preparation.}
To assess whether variable recognition contributes to the performance gap, we evaluate the ability to identify icon-based variables independently of counting. This task isolates visual recognition from symbolic reasoning.

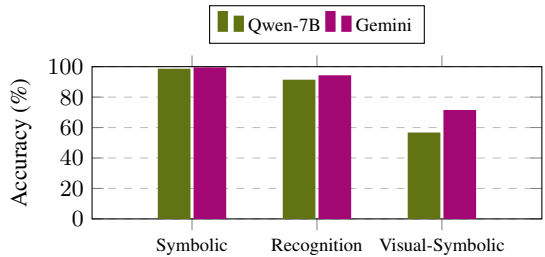
\begin{figure}[!ht]
%\begin{wrapfigure}[12]{r}{0.6\textwidth}
    \centering
    \scalefont{0.8}
    \begin{tikzpicture}
    \begin{axis}[
        ybar,
        bar width=0.4cm,
        width=7.5cm, height=3.6cm,
        enlarge x limits=0.4,
        ylabel={Accuracy (\%)},
        xlabel near ticks,
        ylabel near ticks,
        ymin=0, ymax=100,
        xtick=data,
        %bar shift=0
        xticklabel style={font=\scriptsize},
        symbolic x coords={\textbf{Overall}, Symbolic, Recognition, Visual-Symbolic},
        legend style={at={(0.5,1.15)}, anchor=south, legend columns=3, nodes={scale=0.8, transform shape}},
        ymajorgrids=true,
        grid style=dashed,
    ]

    % Real
    \addplot+[line width=0.3mm, color=ETHGreen] coordinates {
        (Symbolic, 97.9)
        (Recognition, 90.7)
        (Visual-Symbolic, 56.0)
    };
    \addlegendentry{Qwen-7B}
    % Real
    \addplot+[line width=0.3mm, color=ETHPurple] coordinates {
        (Symbolic, 98.8)
        (Recognition, 93.6)
        (Visual-Symbolic, 70.8)
    };
    \addlegendentry{Gemini}
    \end{axis}
    \end{tikzpicture}
    \caption{Accuracy on variable coefficient counting. Results show that both Qwen-7B and Gemini (under the CoT prompt) have difficulty to count the correct value of coefficients.}
    \label{fig:results:recognition}
\end{figure}
%\end{wrapfigure}
\paragraph{Results and Analysis.}
\cref{fig:results:recognition} shows that both Qwen-7B and Gemini achieve high accuracy in recognizing variables from icons (with accuracy above 90\%), with performance comparable to symbolic settings. Details of prompt design are in \cref{app:details:prompt}. This indicates that recognition itself is not a major limitation. Instead, the remaining gap between symbolic and visual-symbolic settings is likely due to task composition, i.e., the challenge of integrating recognition with downstream reasoning.

% \subsubsection{Can VLMs Count?}
\subsubsection{Variable Counting}
\label{app:results:counting}

\begin{figure}[!ht]
%\begin{wrapfigure}[12]{r}{0.6\textwidth}
    \centering
    \scalefont{0.8}
    \begin{tikzpicture}
    \begin{axis}[
        ybar,
        bar width=0.25cm,
        width=7.5cm, height=3.6cm,
        enlarge x limits=0.1,
        ylabel={Accuracy (\%)},
        xlabel near ticks,
        ylabel near ticks,
        ymin=0, ymax=100,
        xtick=data,
        %bar shift=0
        xticklabel style={rotate=20, anchor=east, align=left, font=\scriptsize},
        symbolic x coords={\textbf{Overall}, Qwen-3B, Qwen-7B, Qwen-32B, Qwen-72B, GPT-4o, Gemini},
        legend style={at={(0.5,1.15)}, anchor=south, legend columns=3, nodes={scale=0.8, transform shape}},
        ymajorgrids=true,
        grid style=dashed,
    ]

    % Real
    \addplot+[line width=0.3mm, color=ETHRed] coordinates {
        (\textbf{Overall}, 33.10)
        (Qwen-3B, 32)
        (Qwen-7B, 36.7)
        (Qwen-32B, 24.7)
        (Qwen-72B, 34.6)
        (GPT-4o, 35.14)
        (Gemini, 36.24)
    };
    \addlegendentry{Direct Prompt}
    % Real
    \addplot+[line width=0.3mm, color=ETHBlue] coordinates {
        (\textbf{Overall}, 33.95)
        (Qwen-3B, 30.5)
        (Qwen-7B, 35.6)
        (Qwen-32B, 25.3)
        (Qwen-72B, 35.7)
        (GPT-4o, 40.1)
        (Gemini, 35.6)
    };
    \addlegendentry{CoT Prompt}
    \end{axis}
    \end{tikzpicture}
    \caption{Performance of VLMs on variable counting. Results show that all models have difficulty counting the correct number of variables corresponding to the coefficients.}
    \label{fig:results:counting}
\end{figure}
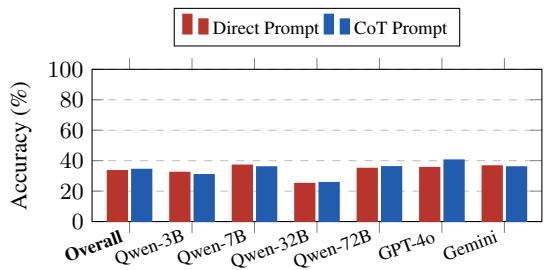
%\end{wrapfigure}
%\cref{fig:results:counting} shows the results of directly evaluating counting ability in the context of equation solving. The corresponding prompt and template design are provided in \cref{app:details:prompt}. These results confirm that counting is the primary bottleneck limiting model performance.
\cref{fig:results:counting} shows the results of directly evaluating counting ability in the context of equation solving. In this setting, models are required to determine the coefficient of a variable by counting the number of repeated object icons. The corresponding prompt and template design are provided in \cref{app:details:prompt}. As shown, all VLMs struggle significantly with this task under both direct and CoT prompting. Although CoT prompting provides a noticeable improvement across all models, the absolute performance remains far below acceptable levels, especially for the smaller open-source models. Notably, even advanced API-based models like GPT-4o and Gemini fail to reach consistent accuracy. This suggests that despite having strong recognition and reasoning abilities in isolation, VLMs are not yet capable of reliably counting visual instances, a key skill required for grounded symbolic reasoning. These results confirm that counting is the primary bottleneck limiting model performance on visual equation solving tasks.

We further investigated the correlation between object quantity and counting accuracy. Test cases were grouped by result ranges (corresponding to the number of icons to be counted), and accuracy was computed for each range. As shown in Table~\ref{tab:counting_accuracy}, counting accuracy declines as the number of icons increases. The Pearson correlation between result value and accuracy is –0.90, indicating a strong negative correlation. These findings indicate that VLMs encounter greater difficulty with higher-count visual inputs, underscoring counting as a fundamental bottleneck for VLMs.

\begin{table}[h]
\centering
\small
\begin{tabular}{lccc}
\toprule
\textbf{Result Range} & \textbf{Total Examples} & \textbf{Accuracy (\%)} \\
\midrule
2--5   & 1,476 & 74 \\
6--10  & 4,452 & 40 \\
11--15 & 4,835 & 20 \\
16--20 & 1,235 & 9 \\
\bottomrule
\end{tabular}
\caption{Counting accuracy across different result ranges.}
\label{tab:counting_accuracy}
\end{table}

\subsection{Three-Variable Equation}
To assess the limitations of VLMs under increased mathematical complexity, we extend our evaluation to systems of three linear equations with \textit{three variables}, which demand more advanced symbolic reasoning and variable tracking than the simpler two-variable case.

\paragraph{Experiment Preparation.}
We generate equations in the same formats as in the default setting: symbolic, visual-symbolic, and fully visual. This allows us to assess whether performance degradation stems from visual perception (i.e., recognition and counting) or from limitations in mathematical reasoning. We report the results under the CoT prompt as it achieves better performance.

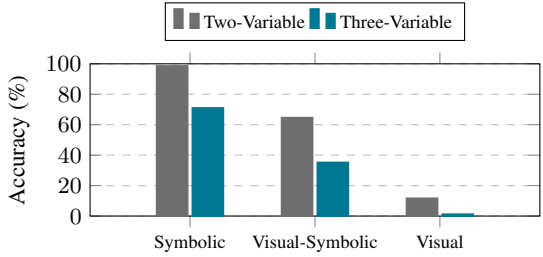
\begin{figure}[!ht]
%\begin{wrapfigure}[12]{r}{0.6\textwidth}
    \centering
    \scalefont{0.8}
    \begin{tikzpicture}
    \begin{axis}[
        ybar,
        bar width=0.4cm,
        width=7.5cm, height=3.6cm,
        enlarge x limits=0.4,
        ylabel={Accuracy (\%)},
        xlabel near ticks,
        ylabel near ticks,
        ymin=0, ymax=100,
        xtick=data,
        %bar shift=0
        xticklabel style={font=\scriptsize},
        symbolic x coords={\textbf{Overall}, Symbolic, Visual-Symbolic, Visual},
        legend style={at={(0.5,1.15)}, anchor=south, legend columns=3, nodes={scale=0.8, transform shape}},
        ymajorgrids=true,
        grid style=dashed,
    ]

    % Real
    \addplot+[line width=0.3mm, color=ETHGray] coordinates {
        (Symbolic, 98.42)
        (Visual-Symbolic, 64.45)
        (Visual, 11.47)
    };
    \addlegendentry{Two-Variable}
    % Real
    \addplot+[line width=0.3mm, color=ETHPetrol] coordinates {
        (Symbolic, 70.83)
        (Visual-Symbolic, 35)
        (Visual, 1)
    };
    \addlegendentry{Three-Variable}
    \end{axis}
    \end{tikzpicture}
    \caption{Overall accuracy across 6 models on solving equations with 2 and 3 variables. Results show that the bottleneck shift from vision side to the math reasoning.}
    \label{fig:results:threevars}
\end{figure}
%\end{wrapfigure}

\paragraph{Results and Analysis.}
As shown in \cref{fig:results:threevars}, model performance drops significantly when moving from two-variable to three-variable systems (accuracy drops from 98\% to 70\% for the symbolic setting, and from 64\% to 35\% for the visual-symbolic setting).
While the visual bottleneck remains largely unchanged, the additional complexity leads to a clear decline in symbolic reasoning. This indicates that, beyond perceptual limitations, current VLMs lack robust mathematical capabilities to solve more complex equation systems.

\subsection{Takeaways.}
Our experiments results show that VLMs perform well on symbolic equations but consistently fail on visual ones. The main bottleneck is visual counting, while variable recognition is largely accurate. However, composing recognition with reasoning introduces significant errors. As equation complexity increases, even symbolic reasoning begins to falter, revealing limits in the models' understanding.

\section{Related Work}
Most existing benchmarks for evaluating VLMs treat perception and reasoning as separate capabilities, rather than testing them as a sequential, integrated process. Recognition-focused datasets such as VQA~\citep{antol-etal-2015-vqa}, GQA~\citep{hudson2019gqa}, and CLEVR~\citep{johnson2017clevr} involve only minimal or trivial arithmetic, which current vision backbones can typically solve with ease. More recent efforts like MathVista~\citep{lu2024mathvista} and DynaMath~\citep{zou2024dynamath} introduce a wider range of visual math problems, but they do not specifically evaluate whether models can solve algebraic equations where symbolic variables and coefficients are visually embedded. The ability to ground a visual system of equations and perform multi-step reasoning over visual cues remains largely untested.

\section{Discussion}
This paper investigates the reasoning limitations of VLMs through visual equation solving, a task that requires combining perception, counting, and symbolic computation. 
While VLMs perform well on symbolic equations and can reliably recognize visual variables, they fail when coefficients must be inferred from repeated visual instances. Our analysis identifies counting and ability composition as key bottlenecks, with performance degrading further as equation complexity increases.

These results highlight gaps in both visual grounding and symbolic reasoning. Addressing them may require new training objectives, compositional architectures, or integration with external tools. Our benchmark provides a diagnostic lens for understanding and improving VLMs on grounded, multi-step reasoning tasks.

% \newpage

\section*{Limitations}
While our study provides insights into the mathematical reasoning capabilities of VLMs, it is subject to a few limitations. 
First, our evaluation focuses primarily on linear equations with integer solutions and addition-only operators. Although this setup allows controlled analysis, it does not capture the full spectrum of mathematical reasoning, such as non-linear or multi-operator problems. 
%Second, our experiments are constrained to two- and three-variable systems; scaling to larger or more abstract equation structures may reveal additional failure modes not captured in our analysis. 
Second, while we isolate key sub-skills like counting and recognition, our diagnostic tasks are still synthetic and could not fully reflect real-world scenarios involving noisy or diverse visual contexts. 
Finally, we rely on prompting-based evaluation, which may under-represent the full potential of models fine-tuned for structured reasoning or equipped with external tools.

\section*{Acknowledgment}
This project was supported by an ETH AI Center Doctoral Fellowship to Junling Wang, the Swiss AI Initiative’s Call for Small Projects (No. 63) to Junling Wang, a Swiss Data Science Center PhD Grant (P22-05) to Yifan Hou, and partial support from the ETH Zurich Foundation. The authors also thank the reviewers for their constructive feedback and the members of the LRE Lab at ETH Zurich.

\bibliography{custom}

\appendix

\onecolumn

\section{Details of Experiment Settings}

\subsection{Data License}
All data used in this study is released under the CC BY 4.0 license. Each generated image is paired with a corresponding question that involves solving one or more equations, along with the ground-truth answers. Users are free to share, adapt, and build upon the dataset, provided appropriate credit is given.

\subsection{Icon List}
\label{app:details:icon}
All the 28 icons that we use are listed below. For each icon, we use only one image to denote the object. Specifically, we select 28 icons labels randomly from the IconQA dataset, and for each label we randomly select one image icon. The icons are: 
\textit{apple}, \textit{palm\_tree}, \textit{strawberry}, \textit{egg}, \textit{clover}, \textit{donut}, \textit{mushroom}, \textit{acorn}, \textit{lemon}, \textit{football}, \textit{flower}, \textit{sheep}, \textit{panda}, \textit{muffin}, \textit{apricot}, \textit{eggplant}, \textit{broccoli}, \textit{rabbit}, \textit{banana}, \textit{rubber\_duck}, \textit{horse}, \textit{fish}, \textit{tomato}, \textit{candy}, \textit{ice\_cream\_cone}, \textit{cake}, \textit{orange}, \textit{carrot}.

\subsection{Model Usage}
\label{app:details:settings}
We conduct inference using four NVIDIA H100 GPUs for each open-source VLM, including Qwen-3B, Qwen-7B, Qwen-32B, and Qwen-72B. All models are loaded using Hugging Face’s Transformers library with automatic mixed-precision (torch.float16 or bfloat16) and memory-efficient device\_map=``auto'' configurations. For each model, we adopt a consistent prompting strategy that combines images and text within structured chat templates. Inputs are tokenized and batched via model-specific processors. Inference is performed on individual image-equation instances using a maximum token length of 2048. We evaluate model outputs using an exact match criterion, comparing extracted variable assignments against ground-truth coefficients. To ensure fairness, we avoid prompt tuning and caching, and run each model independently on the same test set with uniform I/O and decoding procedures. Inference time ranges from 6 to 28 hours depending on model size, with Qwen-72B requiring the longest runtime.

\subsection{Prompt}
\label{app:details:prompt}
\begin{figure}[!ht]
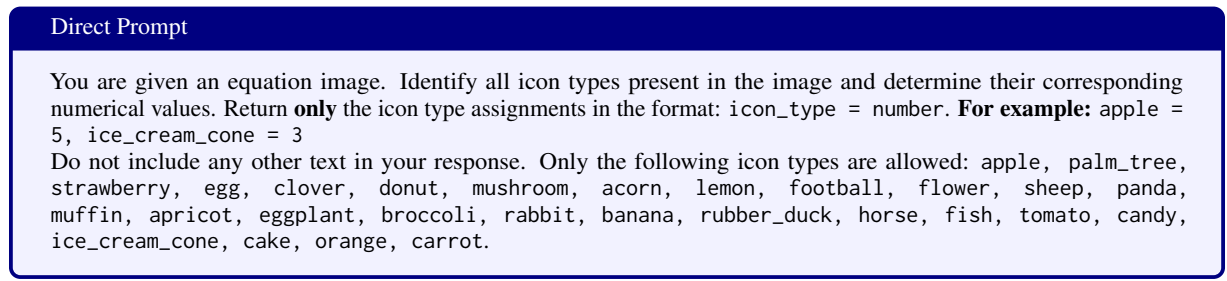

\centering
\small
\begin{tcolorbox}[colback=blue!5!white, colframe=blue!50!black, title=Direct Prompt]
You are given an equation image. Identify all icon types present in the image and determine their corresponding numerical values. Return \textbf{only} the icon type assignments in the format: \texttt{icon\_type = number}. 
\textbf{For example:}
\texttt{apple = 5, ice\_cream\_cone = 3}

Do not include any other text in your response. Only the following icon types are allowed: \texttt{apple, palm\_tree, strawberry, egg, clover, donut, mushroom, acorn, lemon, football, flower, sheep, panda, muffin, apricot, eggplant, broccoli, rabbit, banana, rubber\_duck, horse, fish, tomato, candy, ice\_cream\_cone, cake, orange, carrot}.
\end{tcolorbox}
\caption{\textbf{Direct Prompting Template and Example.} The same prompt is used across all models for consistency.}
\label{fig:zero-shot-prompt}
\end{figure}

\paragraph{Direct Prompting.}
The direct prompt expects models to produce structured outputs in a single step. In our experiments, omitting object labels from the prompt led to poor generation quality, whereas including them significantly improved the reliability and evaluability of the outputs. The prompt template and an example are shown in \cref{fig:zero-shot-prompt}.

\begin{figure*}[!ht]
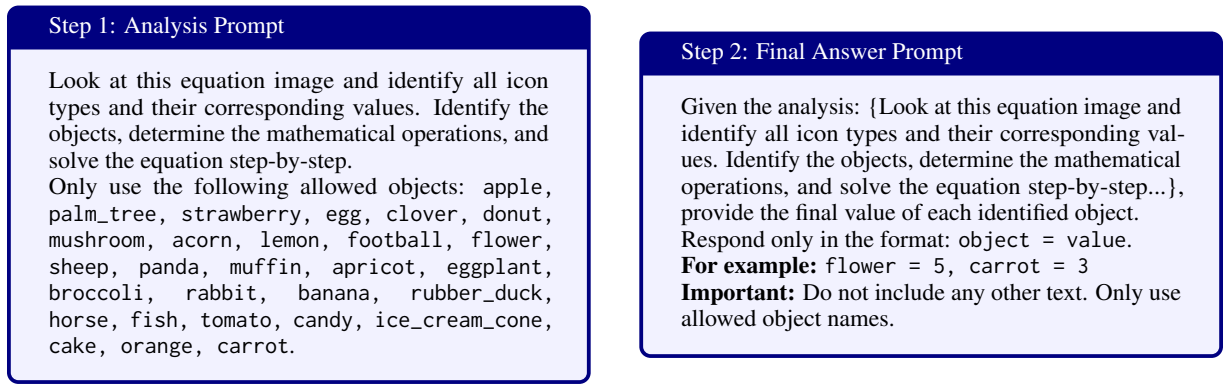

\small
\noindent
\begin{minipage}[h]{0.48\textwidth}
\begin{tcolorbox}[colback=blue!5!white, colframe=blue!50!black, title=Step 1: Analysis Prompt]
Look at this equation image and identify all icon types and their corresponding values. 
Identify the objects, determine the mathematical operations, and solve the equation step-by-step.

Only use the following allowed objects: 
\texttt{apple, palm\_tree, strawberry, egg, clover, donut, mushroom, acorn, lemon, football, flower, sheep, panda, muffin, apricot, eggplant, broccoli, rabbit, banana, rubber\_duck, horse, fish, tomato, candy, ice\_cream\_cone, cake, orange, carrot}.
\end{tcolorbox}
\end{minipage}
\hfill
\begin{minipage}[h]{0.48\textwidth}
\begin{tcolorbox}[colback=blue!5!white, colframe=blue!50!black, title=Step 2: Final Answer Prompt]
Given the analysis: \{Look at this equation image and identify all icon types and their corresponding values. Identify the objects, determine the mathematical operations, and solve the equation step-by-step...\}, provide the final value of each identified object.

Respond only in the format: \texttt{object = value}.\\
\textbf{For example:} \texttt{flower = 5, carrot = 3}

\textbf{Important:} Do not include any other text. Only use allowed object names.
\end{tcolorbox}
\end{minipage}
\caption{\textbf{CoT Prompting Strategy.} The left box initiates free-form reasoning, while the right box extracts the final answers based on the initial prompt and generated response.}
\label{fig:two-step-prompt}
\end{figure*}

\paragraph{Two-Step CoT Prompting.}
To encourage deeper reasoning while avoiding overly rigid output structures, we adopt a two-step chain-of-thought (CoT) prompting strategy. In the first turn, the model is prompted to freely analyze and solve the problem in its own words. In the second turn, we provide both the original prompt and the model’s response, and ask it to extract the final answer. This separation between reasoning and answer extraction allows the model to engage in more flexible, interpretable analysis before committing to a structured output. The prompt used for the object-encoded benchmark is shown in~\cref{fig:two-step-prompt}.

\begin{figure}[!ht]
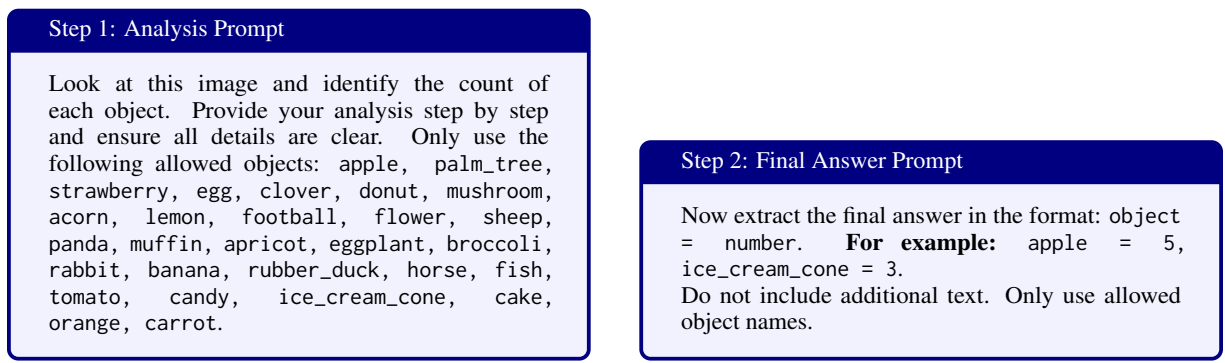

\small
\centering
% Top: Image spanning both prompt columns
%\includegraphics[width=\linewidth]{./figures/counting-example.png}
%\vspace{0.8em}

% Bottom: Two prompt boxes side-by-side
\begin{minipage}[t]{0.48\textwidth}
\begin{tcolorbox}[colback=blue!5!white, colframe=blue!50!black, title=Step 1: Analysis Prompt]
Look at this image and identify the count of each object.  
Provide your analysis step by step and ensure all details are clear.  
Only use the following allowed objects: \texttt{apple, palm\_tree, strawberry, egg, clover, donut, mushroom, acorn, lemon, football, flower, sheep, panda, muffin, apricot, eggplant, broccoli, rabbit, banana, rubber\_duck, horse, fish, tomato, candy, ice\_cream\_cone, cake, orange, carrot}.
\end{tcolorbox}
\end{minipage}
\hfill
\begin{minipage}[t]{0.48\textwidth}
\begin{tcolorbox}[colback=blue!5!white, colframe=blue!50!black, title=Step 2: Final Answer Prompt]
Now extract the final answer in the format: \texttt{object = number}.  
\textbf{For example:} \texttt{apple = 5, ice\_cream\_cone = 3}.  

Do not include additional text. Only use allowed object names.
\end{tcolorbox}
\end{minipage}

\caption{Two-step prompting strategy for solving visual object counting task.}
\label{fig:two-step-counting}
\end{figure}

\paragraph{Counting Prompting (CoT).}
An example input and prompt used for two-step prompting is shown in \cref{fig:two-step-counting}. This prompt is adapted from the CoT strategy and tailored for counting questions involving a single equation, rather than a full system of equations.

%%%%%%
\begin{figure}[!ht]
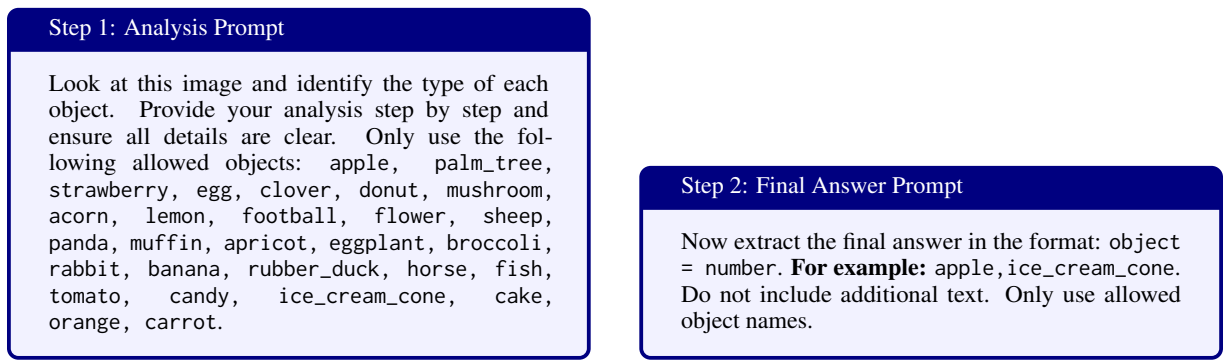

\small
\centering
\begin{minipage}[t]{0.48\textwidth}
\begin{tcolorbox}[colback=blue!5!white, colframe=blue!50!black, title=Step 1: Analysis Prompt]
Look at this image and identify the type of each object.  
Provide your analysis step by step and ensure all details are clear.  
Only use the following allowed objects: \texttt{apple, palm\_tree, strawberry, egg, clover, donut, mushroom, acorn, lemon, football, flower, sheep, panda, muffin, apricot, eggplant, broccoli, rabbit, banana, rubber\_duck, horse, fish, tomato, candy, ice\_cream\_cone, cake, orange, carrot}.
\end{tcolorbox}
\end{minipage}
\hfill
\begin{minipage}[t]{0.48\textwidth}
\begin{tcolorbox}[colback=blue!5!white, colframe=blue!50!black, title=Step 2: Final Answer Prompt]
Now extract the final answer in the format: \texttt{object = number}.  
\textbf{For example:} \texttt{apple,ice\_cream\_cone}.  

Do not include additional text. Only use allowed object names.
\end{tcolorbox}
\end{minipage}

\caption{Two-step prompting strategy for the object-type recognition task.}
\label{fig:two-step-recognition}
\end{figure}

%However, this is in accordance with \cite{deitke2024molmo} and \cite{hou2024do} that discusses benchmarks for counting accuracy for a single-variable visual counting task format and spatial and relational understanding of various entities respectively. For our case, if we assume the object counting process to be independent for each type then it would converge to values which we have obtained in our extended benchmark as seen in Table. \ref{tab:counting-results}. This helps us affirm the hypothesis that the visual module and the multimodal representations that the current VLMs use to interact with language tokens are not robust enough to transfer the entire visual-count information correctly leading to the entire chain-of-reasoning to collapse and eventually lead to a wrong answer even when the solving steps are correct on their own. 

\paragraph{Recognition Prompting (CoT).}
% {\color{red} TODOs: please paste the prompt template and examples similarly to the counting.}

An example input and prompt used for two-step prompting is shown in \cref{fig:two-step-counting}. This prompt is adapted from the CoT strategy and tailored for recognizing the object-type present in the image. We utilize the same single equation dataset as mentioned for the counting task. 

\subsection{Data Collection and Anonymization Procedures}
To ensure ethical use and privacy compliance, we carefully selected data from publicly available, open-access visual datasets that contain no personally identifiable information or sensitive content. Specifically, we constructed our benchmark using synthetic images generated from programmatically defined math equations and object icons sourced from the IconQA dataset~\citep{iconqa}. These icons depict generic, non-human items such as apples, bananas, and lemons, eliminating the risk of capturing real-world identities or offensive imagery. Furthermore, we manually reviewed a sample of the generated images to confirm the absence of inappropriate or sensitive elements.
\section{Supplementary Results}

\subsection{Case Study}
\label{app:results:case_study}

\begin{figure}[ht!]
\small
\centering

\includegraphics[width=0.8\textwidth]{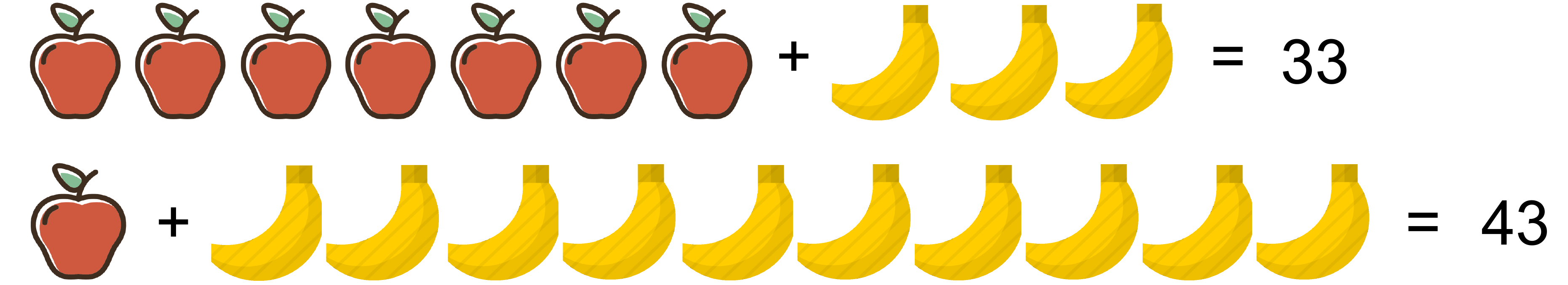}  % Adjust filename and width as needed

\vspace{1em}

\begin{tcolorbox}[title=Gemini 2.0 Flash Response,colframe=blue!50!black,colback=blue!5!white, width=\textwidth]

\noindent\textbf{Step 1: Object Identification}

\noindent apple = \(x\) \\
banana = \(y\)

\vspace{1em}
\noindent\textbf{Step 2: Operation Analysis} \textcolor{red}{(Wrong Counting)}

\[
\begin{aligned}
6x + 3y &= 33 \\
x + 9y &= 43
\end{aligned}
\biggl\} \quad
\begin{aligned}
\textcolor{red}{\text{Wrong count! The original equations are}} \\
\textcolor{red}{7x+3y=33} \\
\textcolor{red}{1x+10y=43}
\end{aligned}
\]

\vspace{1em}
\noindent\textbf{Step 3: Calculation}

Substitute into second equation:

\[
x = \frac{33 - 3y}{6}
\]

\[
\frac{33 - 3y}{6} + 9y = 43 \Rightarrow 33 - 3y + 54y = 258 \Rightarrow 51y = 225 \Rightarrow y = \frac{225}{51} = \frac{75}{17} \approx 4.41
\]

Now, solve for \(x\):

\[
x = \frac{33 - 3 \cdot \frac{75}{17}}{6} = \frac{33 - \frac{225}{17}}{6} = \frac{\frac{561 - 225}{17}}{6} = \frac{336}{17 \cdot 6} = \frac{56}{17} \approx 3.29
\]

Trial with integer values:

\[
y = 5 \Rightarrow x = \frac{33 - 15}{6} = 3
\]

Check:

\[
6 \cdot 3 + 3 \cdot 5 = 33 \quad (\checkmark) \quad 3 + 9 \cdot 5 = 48 \quad (\textcolor{red}{\text{\sffamily X}})
\]

Alternative derivation by elimination:

\[
6x + 54y = 258 \quad \text{(eq2 × 6)} \\
(6x + 54y) - (6x + 3y) = 258 - 33 \Rightarrow 51y = 225 \Rightarrow y = \frac{75}{17}
\]

\[
x = 43 - 9 \cdot \frac{75}{17} = \frac{731 - 675}{17} = \frac{56}{17}
\]

\vspace{1em}
\noindent\textbf{Step 4: Final Answer}

\[
\boxed{\text{apple} = \frac{56}{17} \quad \text{banana} = \frac{75}{17}}
\]

\end{tcolorbox}
\caption{Response of Gemini 2.0 Flash for the given query image (top) using the CoT prompting. Note the mistake made by the model in coefficient counting step (bottom). This cascades to the model eventually giving the wrong answer. The correct answer to this problem is apple ($x$) = 3, banana ($y$) = 4. }
\label{fig:wrong-example}
\end{figure}

An example of an intermediate generation output is shown in \cref{fig:wrong-example}. Upon closer inspection, we find that in most failure cases, the model incorrectly determines the coefficients during equation interpretation. 
%Since coefficients in our dataset correspond to object counts, this raises two key questions: Can VLMs accurately recognize and count visual objects? And if so, can they combine these abilities with arithmetic reasoning? A deeper analysis of these aspects helps address the core question driving our evaluation: Are VLMs reasoning in the blind?

\subsection{Experiments with Reasoning Models}
To validate whether reasoning models share the same limitations as standard VLMs, we conducted experiments with o4-mini and Gemini 2.5 Pro. Since reasoning models already possess chain-of-thought abilities, we only used direct prompting with them. The results are shown in Table~\ref{tab:visual_equation}. These results are consistent with our original findings: VLMs perform well in visual recognition, symbolic equation solving, and visual-symbolic equation solving, but struggle with visual counting and visual equation solving. This further supports our core conclusion that counting and ability composition remain key bottlenecks for current VLMs. 

\begin{table}[h]
\centering
\scriptsize
\begin{tabular}{l p{2cm} p{2.8cm} p{2.3cm} p{1.7cm} p{2cm}}
\toprule
\textbf{Model} & \textbf{Visual Equation Solving (\%)} & \textbf{Visual-Symbolic Equation Solving (\%)} & \textbf{Symbolic Equation Solving (\%)} & \textbf{Visual Counting (\%)} & \textbf{Visual Recognition (\%)} \\
\midrule
o4-mini        & 36.5 & 90.4  & 98.9  & 41.0 & 90.1 \\
Gemini 2.5 Pro & 43.1 & 91.3  & 98.2 & 61.3 & 91.3 \\
\bottomrule
\end{tabular}
\caption{Performance comparison across different visual and symbolic tasks.}
\label{tab:visual_equation}
\end{table}

\section{Potential Risk}
Our study involves the generation of synthetic visual math equations using object icons, and evaluation is conducted using publicly available open-source models and commercially accessible API-based VLMs. As our work does not involve real-world data, human subjects, or sensitive content, we do not anticipate any ethical concerns or foreseeable risks associated with this research.

\section{Use of AI Assistants in Research}
In our study, AI assistants were used sparingly and in accordance with ACL's Policy on AI Writing Assistance. We utilized ChatGPT and Grammarly for basic paraphrasing and grammar checks, respectively. These tools were applied minimally to ensure the authenticity of our work and to adhere strictly to the regulatory standards set by ACL. Our use of these AI tools was focused, responsible, and aimed at supplementing rather than replacing human input and expertise in our research process.

\end{document}